\documentclass[conference]{IEEEtran}
\IEEEoverridecommandlockouts
\usepackage{cite}
\usepackage{booktabs}
\usepackage{makecell}
\usepackage{multirow}
\usepackage{amsthm,amsmath,amssymb,amsfonts}
\theoremstyle{definition}
\newtheorem{definition}{Definition}
\usepackage{algorithmic}
\usepackage{graphicx}
\usepackage{textcomp}
\usepackage{xcolor}
\usepackage{fancyhdr}
\def\BibTeX{{\rm B\kern-.05em{\sc i\kern-.025em b}\kern-.08em
    T\kern-.1667em\lower.7ex\hbox{E}\kern-.125emX}}

\DeclareRobustCommand*{\IEEEauthorrefmark}[1]{%
  \raisebox{0pt}[0pt][0pt]{\textsuperscript{\footnotesize #1}}%
}

\begin{document}

\title{Generative AI-based Prompt Evolution Engineering Design Optimization With Vision-Language Model}

\author{
\IEEEauthorblockN{Melvin Wong\IEEEauthorrefmark{1}, 
Thiago Rios\IEEEauthorrefmark{2},
Stefan Menzel\IEEEauthorrefmark{2},
Yew Soon Ong\IEEEauthorrefmark{1}} \\ \vspace{-3mm}
\IEEEauthorblockA{ 
\textit{\IEEEauthorrefmark{1}College of Computing \& Data Science (CCDS), Nanyang Technological University (NTU), Singapore} \\
\textit{\IEEEauthorrefmark{2}Honda Research Institute Europe (HRI-EU), Offenbach am Main, Germany} \\ \vspace{-3mm} \\
\{wong1357, asysong\}@ntu.edu.sg, \{thiago.rios, stefan.menzel\}@honda-ri.de
}}

\maketitle

\fancypagestyle{footmark}{
  \fancyhf{}
  \fancyfoot[L]{\footmark}
  \renewcommand{\headrulewidth}{0pt}
  \renewcommand{\footrulewidth}{0.4pt}
}
\newcommand\markfoot[1]{\gdef\footmark{#1}\thispagestyle{footmark}}

\markfoot{\footnotesize \textcopyright 2024 IEEE. Personal use of this material is permitted.  Permission from IEEE must be obtained for all other uses, in any current or future media, including reprinting/republishing this material for advertising or promotional purposes, creating new collective works, for resale or redistribution to servers or lists, or reuse of any copyrighted component of this work in other works.}

\begin{abstract}
Engineering design optimization requires an efficient combination of a 3D shape representation, an optimization algorithm, and a design performance evaluation method, which is often computationally expensive. We present a prompt evolution design optimization (PEDO) framework contextualized
in a vehicle design scenario that leverages a vision-language model for penalizing impractical car designs synthesized by a generative model. The backbone of our framework is an evolutionary strategy coupled with an optimization objective function that comprises a physics-based solver and a vision-language model for practical or functional guidance in the generated car designs. In the prompt evolutionary search, the optimizer iteratively generates a population of text prompts,
which embed user specifications on the aerodynamic performance and visual preferences of the 3D car designs. Then, in addition to the computational fluid dynamics simulations, the pre-trained
vision-language model is used to penalize impractical designs and, thus, foster the evolutionary algorithm to seek more viable designs. Our investigations on a car design optimization problem
show a wide spread of potential car designs generated at the early phase of the search, which indicates a good diversity of designs in the initial populations, and an increase of over 20\% in the probability of generating practical designs compared to a baseline framework without using a vision-language model.
Visual inspection of the designs against the performance results demonstrates prompt evolution as a very promising paradigm for finding novel designs with good optimization performance while providing ease of use in specifying design specifications and preferences via a natural language interface.
\end{abstract}

\vspace{-2mm}
\section{Introduction}
\vspace{-1mm}
Generative artificial intelligence (GenAI) and large language models (LLMs) are currently contributing to major advances in a variety of fields, such as text and image-based content generation, dialog-based systems, knowledge retrieval and management, and co-programming, among others.
The intuitive interaction through natural language processing (NLP) using text prompts allows non-experts to access domain knowledge and execute domain-specific processes in a cooperative fashion. 
In the field of product development, human users strive for the generation of novel and realizable designs, \textit{i.e.}, images, and 3D shapes, which fulfill pre-defined specifications and ideally perform well in a variety of environment conditions.
For image or video synthesis from text prompts, there exists a portfolio of tools, \textit{e.g.} OpenAI's Dall-E 3 \cite{Betker2023}, stable diffusion \cite{rombach2021highresolution}, Midjourney and Lumiere \cite{bartal2024lumiere}, which provides high-quality visual results taken into account user ideas, constraints, style requests, and further context.

Differently, 3D shape synthesis through prompt-based data-driven generative models is still very challenging.
In the field of geometric deep learning \cite{Bronstein2017}, there are several successful attempts to utilize generative models, such as autoencoders, to learn 3D design data and perform geometric operations based on the learned latent representations \cite{Achlioptas2018, Umetani2017, Gao2019}.
However, while image-generative models rely on a massive amount of (annotated) data, the number of publically available annotated data sets containing 3D objects represented as watertight and non-skewed meshes for training text-to-3D models is much lower.
In addition, generating 3D design data with high variability and corresponding performance as annotation, \textit{e.g.} vehicle aerodynamic drag, requires high manual and computational effort, which hinders the development of new data sets for GenAI applications in industrial settings.
Despite the challenges, with the advance of LLMs, OpenAI recently released Point-E \cite{Nichol2022} and Shap-E \cite{Jun2023}, which allow users to generate 3D shapes from text prompts.

\begin{figure}[ht]
    \centering
    \begin{tabular}{cc}    
     \includegraphics[width=35mm]{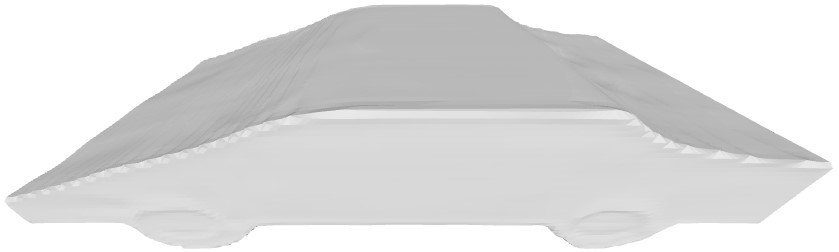} &
     \includegraphics[width=35mm]{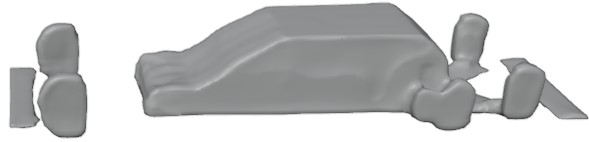} \\ 
     (a) Practical Car Design & (b) Impractical Car Design \\[6pt]
    \end{tabular}
    \caption{Text-to-3D model, Shape-E \cite{Jun2023}, using the prompt "a fast car in the shape of a wing" can synthesize practical novel car designs (Fig. \ref{fig:compare_feasibility} (a)). However, it sometimes ``hallucinates'', resulting in a car design (Fig. \ref{fig:compare_feasibility} (b)) that still resembles a ``car'' but bears a malformed shape and has disconnected geometries, rendering such design impractical for engineering design.}
    \label{fig:compare_feasibility}
    \vspace{-2mm}
\end{figure}

The ease of using text-to-\emph{X} generative models as black-box to obtain a diverse set of designs quickly has garnered significant interest in the computation engineering design optimization domain. 
Although the interfaces allow the users to express design specifications in a natural way, the practicality of novel designs synthesized by such generative models needs to be carefully treated and assessed, particularly for the design variations synthesized throughout a design optimization task.
Since generative models learn the distribution of the representation of various objects, it is unclear which variety of shapes can be combined or interpolated to generate practical novel designs. 
Additionally, generative models tend to ``hallucinate'' designs, especially for prompts with multiple concepts or preferences, which potentially fit the description in the input prompt but can be impractical for engineering design.
For example, the prompt ``a fast car in the shape of a wing" should lead a text-to-3D model to generate a vehicle-like geometry that resembles the shape of a wing (Fig. \ref{fig:compare_feasibility} (a)) but some of the generated models might be ill-defined designs that still resemble a car (Fig. \ref{fig:compare_feasibility} (b)). 
Hence, the stochasticity of text-to-\emph{X} models can significantly impact the evolution process of a fully automated evolutionary design optimization framework by leading the global search to regions with many optimal yet impractical designs.

To mitigate the generation of ``hallucinated'' designs, in the present paper, we propose a fully automated prompt evolution design optimization (PREDO) framework based on an evolutionary design optimization process, introduced in \cite{rios2023LLM}, with a pre-trained vision-language model that identifies impractical designs synthesized by generative models.
Through this novel extension, we improve the optimization by penalizing the fitness of impractical designs that are identified by the vision-language model and, thus, have a lower rank than practical designs during scoring and selection.
Hence, due to the clearer detection of viable designs, we expect the deployed evolutionary algorithm to construct prompts more likely to yield practical designs.
In our experiments, we utilize different vision-language models and evaluate our PREDO framework by comparing the optimization performance with respect to corresponding unconstrained cases, which allows us to identify the contribution of the penalization to the objective function and impact of the vision-language model in the optimization process.

In summary, our contributions are as follows:

\begin{itemize}
    \item A prompt evolutionary design optimization (PREDO) framework that incorporates two language-based AI models, (1) a generative text-to-3D model for shape synthesis and (2) a vision-language model to rank practical designs higher than impractical ones, allowing more viable designs to be found.
    \item An optimization objective function that employs a physics solver and a vision-language model for physical and visual guidance, working in harmony during the evolutionary process to find novel and practical designs.
\end{itemize}

The remainder of this paper is organized as follows: In Section \ref{sec:review}, we discuss notable works in the literature related to 3D generative models for design optimization and prompt tuning applied to text-to-\emph{X} models.
Section \ref{sec:framework} provides an overview of our proposed fully automated evolutionary design optimization PREDO framework, highlighting various strategies for constructing the text prompt during the evolution process and detailing how vision-language models can be integrated into the framework.
Section \ref{sec:experiments} presents the application of our proposed method in design optimization for minimizing aerodynamics or projected frontal area of car designs and provides a comprehensive analysis of our experiments.
We conclude this paper in Section \ref{sec:conclusion} and highlight potential future research directions.
\vspace{-2mm}
\section{Literature Review} \label{sec:review}
\vspace{-1mm}
The recent advances in hardware and software technologies, \textit{e.g.} new powerful graphic cards (GPUs) and dedicated libraries for GPU-based computation, enabled the development of more sophisticated machine learning architectures, such as geometric deep learning \cite{Bronstein2017} techniques for training 3D shape-generative models.
In the context of engineering tasks, promising results have been achieved with the learning-based representation of 3D shapes, such as for car aerodynamic design and optimization \cite{Umetani2017, Rios2021c}, as well as surrogate modeling for performance prediction \cite{Wollstadt2022,Saha2021}.
In these cases, as the latent variables are learned from large design data sets, \textit{e.g.} ShapeNetCore \cite{Chang2015}, the latent space potentially provides the optimization algorithm a broader range of solutions than typical design variables selected by a human user.
However, the representations learned by these models lack interpretability, which hinders the application of more specific geometric design constraints in optimization or manual modification of the designs.

Recently, novel text-based generative models have been proposed, which allow the user to specify the design preferences using natural language explicitly.
Point-E \cite{Nichol2022} and Shap-E \cite{Jun2023}, proposed by OpenAI, were the first models that generate 3D objects from text prompts.
While Point-E was an initial model based on 3D point cloud representations, Shap-E provides more advanced features, such as a differentiable implementation of the marching cubes 33 algorithm to generate output shapes as polygonal meshes, which are often required for rendering objects in computer graphics or for simulation-based applications in engineering.

Shap-E was recently integrated into a vehicle aerodynamic drag minimization problem and showed that Shap-E is a suitable generative model for engineering optimization \cite{rios2023LLM}.
In that work, the authors proposed a fully automated evolutionary design optimization framework incorporating evolutionary strategy and performance-based guidance to find novel 3D car designs.
Nevertheless, the authors also claim that the design search in the text-prompt space is still challenging, even more so as Shap-E was not trained specifically on a data set of car shapes, potentially leading to many ill-defined designs.

Lately, some work has been done to generate annotated data sets focused on design modifications through text prompts \cite{Achlioptas2023}, which improves the control over the design modifications but the available data are still limited to a small number of object classes.
However, a more promising approach to improve the performance of design optimizations based on text prompts is to add prompt tuning mechanisms, which avoid the generation of prompts that yield ill-defined meshes or designs, unlike car shapes.


Arechiga et al. train a dedicated surrogate model to provide physics-based guidance into the denoising process of a diffusion model \cite{arechiga2023drag}, opening up new possibilities for optimizing 2D car designs with respect to a performance metric during the generation process. However, such an approach requires an extensive collection of training datasets, which is not applicable in most engineering design optimization scenarios. 

The recently conceptualized prompt evolution paradigm for GenAI \cite{wong2023promptevo} demonstrated the feasibility of leveraging the evolutionary process and a vision-language model as the black-box preference guidance to synthesize multiple outputs that best satisfy target preferences inferred by the text prompt. 
In this work, examples showed how the target preferences are served via a natural language interface to the CLIP model \cite{pmlr-v139-radford21a} for assessing the probability of generated images satisfying the target preferences. 
Using such pre-trained vision-language models \cite{pmlr-v139-radford21a,li2023blip} avoids the need to train a dedicated model and can also provide ease of use in specifying the design specifications and preferences with good generalizability and accuracy.
Hence, in the present work, we combine the benefits of learning-based representations, the prompt evolution paradigm, and visual-based and physics-based guidances to propose a novel framework for solving engineering design optimization problems.

\vspace{-2mm}
\section{Methods} \label{sec:framework}
\vspace{-1mm}
Evolutionary design optimization is widely used in real-world engineering applications where performance objective gradients are inaccessible or incomputable.
In contrast to traditional numerical design encodings, we rely in this paper on a language-model-based representation following \cite{rios2023LLM} and improve the generation of valid car designs by including a vision-language model. To this end, we conceptual the idea that we define as follows:
\vspace{-2mm}
\begin{definition}[Prompt Evolutionary Design Optimization]\label{defevo} 
\emph{Prompt Evolutionary Design Optimization (PREDO) imparts evolutionary variation to the prompt that best satisfy the design objective(s), with the aim of guiding generative AI to synthesize practical designs.
.}
\end{definition}
\vspace{-2mm}
\noindent Our proposed prompt evolutionary design optimization (PREDO) framework comprises four key components: (1) a derivative-free evolutionary optimizer, (2) a text-to-3D generative model, (3) a novel integration of vision-language models, and (4) a physics-based solver to evaluate car design performance. 

\subsection{Overview}

\begin{figure*}[htb]
    \vspace{-5mm}
    \centering
    \includegraphics[width=1.0\textwidth]{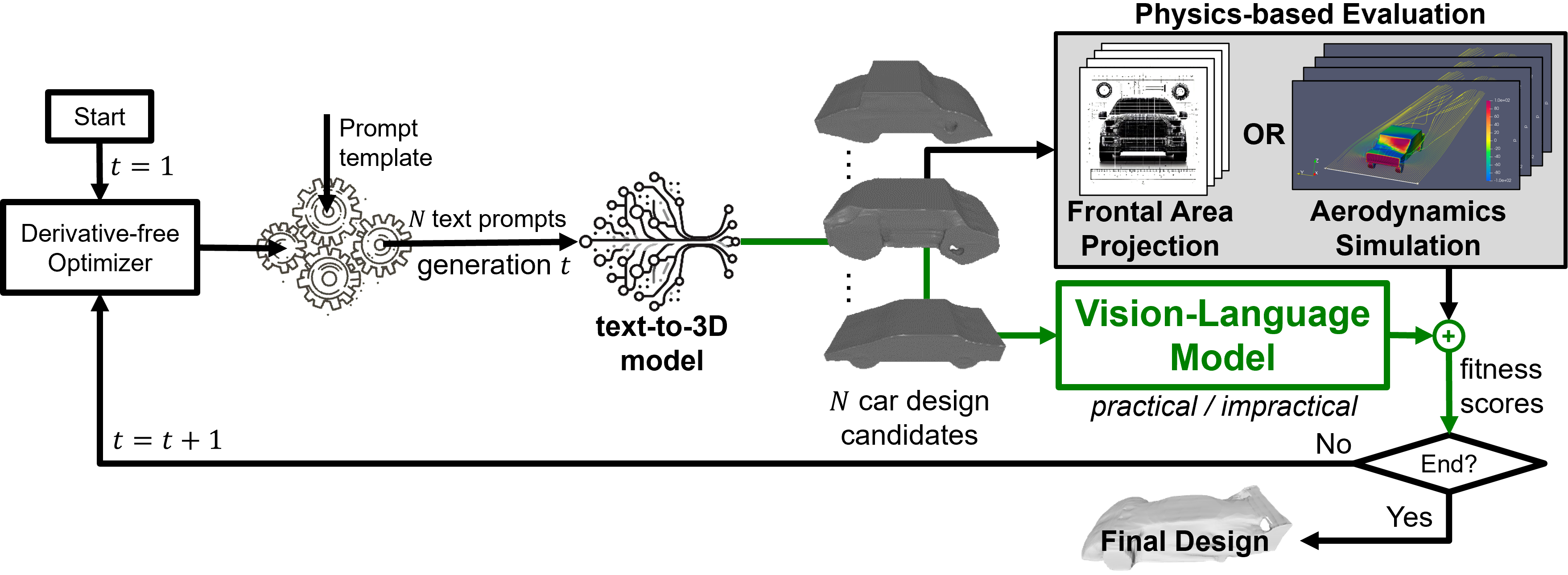}
    \caption{Building upon the fully automated evolutionary optimization framework \cite{rios2023LLM}, the vision-language model is introduced into the prompt evolutionary design optimization (PREDO) framework, guiding the optimizer to find more viable designs.}
    \label{fig:framework}
    \vspace{-5mm}
\end{figure*}

Our framework begins with the optimizer sampling a population of $N$ set of words where each word is selected based on a distance function between a reference word and its corresponding sample word (Fig. \ref{fig:framework}). 
Each set of words is then used to instantiate the prompt template containing the immutable specifications expressed in free-form natural language. 
The $N$ text prompts are provided to the text-to-3D generative model, which synthesizes $N$ designs, with each design represented as 3D polygonal meshes that best satisfy the prompt. 
The synthesized 3D designs are then post-processed for the physics-based solver to evaluate the design performance. 
The design performance $s_{design}$ is then computed and normalized using the following equation to serve as the fitness scores for the optimizer. 

\vspace{-1mm}
\begin{equation} \label{eq:fittness}
    f_{score} = \frac{s_{design}}{max(s_{base}) - min(s_{base)}}
\end{equation}

\noindent Here, $s_{base}$ is the baseline design performance of a reasonably large population of design of interest generated using the text-to-3D generative model.

The design optimization objective is then:
\vspace{-1mm}
\begin{equation} \label{eq:obj}
    \min f_{score}
\end{equation}

Based on the convergence criteria, such as maximum generations or minimal improvement gained, the optimization iteration either terminates and presents the last generation designs for user selection or applies the selected evolutionary selection pressure strategy and repeats the evolution process for the next generation.

\subsection{Strategies for Constructing the Prompt} \label{sec:strategies}

In the framework from the reference work, the authors proposed two strategies for constructing the prompt \cite{rios2023LLM}.
The first strategy employs the Wordnet \cite{wordnet1998} vocabulary set as bag-of-words (BoW), which the optimizer will sample pairs of $<$\emph{adjective}$>$ and $<$\emph{noun}$>$ to instantiate the following prompt template \vspace{-2mm} \\

\centerline{A $<$\emph{adjective}$>$ car in the shape of $<$\emph{noun}$>$.} 

\noindent \vspace{-2mm} \\ The optimizer will select the sample words using Wu \& Palmer (WUP) similarity metric \cite{wu1994verb} with respect to the reference $<$\emph{adjective}$>$ and $<$\emph{noun}$>$ words. 

On the other hand, the second strategy deploys the following prompt template \vspace{-2mm} \\

\centerline{A car in the shape of $<$\emph{string}$>$,}

\noindent \vspace{-2mm} \\ where the $<$\emph{string}$>$ comprises of the most common words in vocabulary. Here, the evolutionary optimizer will sample the common words from a normal distribution.

As an unconstrained optimization problem, the synthetic generation of the prompts may force the model to sample out of the learned distribution and generate ill-defined designs \cite{rios2023LLM}.
Albeit ill-defined designs are impractical in the real world, \textit{e.g.}, flat car designs without space for passengers, some of the geometries yield better aerodynamic performance than the other individuals in the population and are selected as parents to generate new offspring.
Consequently, the optimization can foster the generation of prompts that lead to other impractical high-performance designs that do not meet the user requirements.
Therefore, we propose adding a soft constraint using a vision-language model to avoid generating ill-defined designs during the optimization.

\subsection{Prompt Evolutionary Design Optimization with Vision-Language Model} \label{sec:cpromptevo}

In this section, we propose a technique that addresses the generation of ill-defined designs by a text-to-3D model as candidates in the evolution process, which we seamlessly integrate into the framework discussed in Section \ref{sec:framework} A and B.

\subsubsection{Vision-Language Model as Soft Constraint}
A straightforward way is to employ a black-box solver to identify ill-defined car designs based on the expected design performance range and remove or penalize such designs. However, such a solver is computationally expensive. Furthermore, removing impractical car designs as a post-processing step complicates the mutation and selection pressure processes. In addition to that, a larger population size is required to ensure enough solutions can make it to the selection process. However, synthesizing large batches of designs using text-to-\emph{X} models is also computationally expensive. As such, a better alternative is identifying and penalizing ill-defined car designs based on its visual preferences using a vision-language model. Given a model $g$, we compute the penalty score as follows
\vspace{-1mm}
\begin{equation} \label{eq:penalty}
    s_{penalty} = -log\left(g(\textbf{x}, \textbf{p}_{target})\right),
\end{equation}

\noindent where $\textbf{x}$ is a single viewpoint image of the 3D design synthesized using the text-to-3d model, $\textbf{p}_{target}$ is the encoded text prompt of visual preferences and $s_{penalty} > 0$. Here, we assume the model yields a high degree of practicality (close to one) for designs that satisfy the text prompt and a low degree of practicality score (close to zero) for designs otherwise.

\subsubsection{Engineering Design Optimization}

Without loss of generality, as illustrated in Fig. \ref{fig:framework}, we introduce the penalized term in Equation \ref{eq:fittness} and \ref{eq:obj}, resulting in the following optimization objective
\vspace{-1mm}
\begin{equation}
    \min \hat{f}_{score} = \frac{s_{design}}{max(s_{base}) - min(s_{base})} + \alpha s_{penalty},
\end{equation}

\noindent where $\alpha > 0$ is the penalty weight that controls the contribution of the penalized term. Notice that when the design is deemed viable, $\hat{f}_{score} \approx f_{score}$ as $s_{penalty}$ will be infinitesimal. On the other hand, when the design is identified as impractical visually, $\hat{f}_{score} > f_{score}$. Hence, this optimization objective reorders the generated 3D designs such that practical designs are ranked higher than impractical ones.
\vspace{-2mm}
\section{Experiments and Discussion} \label{sec:experiments}
\vspace{-1mm}
We present our experiments and analysis to demonstrate the performance of the baseline framework and our proposed method. 

\subsection{Experiment Settings}

Following \cite{rios2023LLM}, we employ the Shape-E model \cite{Jun2023} as text-to-3D model, and the derandomized evolution strategy with covariance matrix adaptation (CMA-ES) \cite{hansen2001ecj} as derivative-free evolutionary optimizer with the same hyperparameters for population size $\lambda = 10$, number of parents $\mu = 3$ and a maximum number of generations set to 100. CMA-ES has shown high efficiency for problems with comparably small populations and limited allowed numbers of function evaluations, which is typical for engineering design optimization applications. 

In addition, we obtained the baseline performance of a car design that the Shape-E model will generate based on 300 car designs synthesized by the model using the prompt ``A car". As such, we obtained a similar mean normalized drag coefficient range as observed in \cite{rios2023LLM}. Furthermore, to demonstrate the efficacy of using a vision-language model, we employ two vision-language models to compare the performance, BLIP-2 \cite{li2023blip}, and CLIP \cite{pmlr-v139-radford21a}, set the penalty weight for both models $\alpha = 1.0$. In addition, for consistency, all our experiments use the same random seed. Finally, as highlighted in \cite{rios2023LLM}, the projected frontal area has the most significant impact on the aerodynamics of the car designs. As such, we used this metric as the design performance to compare with the normalized drag coefficient of the car.

We performed the optimization and simulations in parallel on a single shared compute node with two different configurations. One configuration comprises Intel Xeon Silver 64 CPU cores, clocked at 2.10 GHz, 128GB of RAM, and equipped with 3 Nvidia Quadro GV100 GPUs (32 GB each). The other configuration includes AMD Ryzen Threadripper PRO 5965WX with 24 CPU cores, clocked at 2.34 GHz, 512GB of RAM, and equipped with 2 Nvidia RTX A6000 GPUs (48 GB each). The GPUs are used by the Shape-E text-to-3D model and the aforesaid vision-language models to generate new design candidates and assess the feasibility of the design with respect to the text prompt ``a car".

We adopt the following approaches to setup the strategies discussed in Section \ref{sec:strategies} for constructing the prompts:

\subsubsection{Select reference words in BoW strategy} \label{sec:bow_strategy}

We chose the same reference $<$adjective$>$ and $<$noun$>$ words used in \cite{rios2023LLM} experiments, \\

\centerline{A \emph{fast} car in the shape of \emph{wing}.} \hspace{0pt}

These reference words are expected to yield low drag coefficients. Our proposed method computes WUP similarities between these reference and the WordNet words, which are then used to select the WordNet words with the smallest difference between a WUP similarity and the continuous variable provided by the CMA-ES.

\subsubsection{Constructing the $<$string$>$ in the tokenization strategy} \label{sec:ttok_strategy}


We adopt the following approach to select the most common words in vocabulary. Let $\mathcal{X} \subset \mathbb{R}^{D}$ denote a space of latent variables of $D$ dimensions. We use a linear model $h(\textbf{z}): \mathbb{R}^{D} \rightarrow [0,1]^{V}$ to select a token out of $V$ tokens in GPT-4 \cite{achiam2023gpt} vocabulary. Given $N$ population of $M$ length text prompts $\textbf{x}^{[i, j]} \in \mathcal{X}$ where $i \in \{1,...,N\}$ and $j \in \{1,...,M\}$, we sample the tokens in the vocabulary using the following equations:
\vspace{-3mm}
\begin{equation}
    \begin{split}
        \textbf{z}_{[j]} \sim N(\mu, \Sigma),\\
        \textbf{z} \subset \mathbb{R}^{D \times N}
    \end{split}
\end{equation}
\vspace{-3mm}
\begin{equation}
    \begin{split}
        h(\textbf{z}_{[j]}) = \arg \max \left(sigmoid(\textbf{z}_{[j]}^{\textrm{T}}\textbf{W})\right),\\
        \textbf{W} \subset \mathbb{R}^{D \times V},
    \end{split}
\end{equation}

\noindent where $\textbf{W}$ is the weights of the linear model, and $\mu = 3$ and $\Sigma = 0.2$ are the parameters of CMA-ES. Here, constructing $N$ population of text prompts requires initializing the prompt template by first having CMA-ES generate $M$ latent variables that we provide to the linear model $h(\textbf{z})$ to select $M$ tokens for the $<$string$>$.

\subsection{Experiment Results and Discussion}

We evaluate our proposed method against the baseline framework on two strategies discussed in Section \ref{sec:bow_strategy} and \ref{sec:ttok_strategy}. During optimization, we deploy CMA-ES to optimize 3D car designs by evolving the prompt template following the aforementioned approaches. 

\begin{table}[b!]
	\centering
	\vspace{-3mm}
	\caption{Comparing between PREDO variants and its baseline in finding viable car design using the \textbf{Bag-of-Words strategy}}
        \vspace{-1mm}
	\begin{tabular}{lccc}
		\toprule
		\makecell{Design \\ Objective} & Method & \makecell{Gen. 100 \\ Avg. Accuracy} & \makecell{Overall \\ Avg. Acc.$\pm$Std. Dev.} \\
		\midrule
		Projected & Baseline & $30.00\%$ & $28.10 \pm 0.0171\%$ \\
        Frontal & $\textrm{PREDO}_{clip}$ & $\textbf{70.00\%}$ & $\textbf{59.60} \pm \textbf{0.0218\%}$ \\
        Area & $\textrm{PREDO}_{blip2}$ & $50.00\%$ & $58.90 \pm 0.0275\%$ \\
        \midrule
        Normalized & Baseline & $30.00\%$ & $53.00 \pm 0.0245 \%$ \\
        Drag & $\textrm{PREDO}_{clip}$ & $\textbf{60.00\%}$ & $\textbf{59.70} \pm \textbf{0.0205\%}$ \\
        Coefficiency & $\textrm{PREDO}_{blip2}$ & $60.00\%$ & $57.60 \pm 0.0289\%$ \\
        \bottomrule
	\end{tabular}
	\label{tbl:bow}
\end{table}

In the experiments conducted using the projected car frontal area as the design performance objective as presented in Table \ref{tbl:bow} and Table \ref{tbl:tok}, we observed a significant improvement of more than $20\%$ in finding practical car designs when the vision-language model is introduced. Notice that in Table \ref{tbl:bow}, when the design performance objective is switched to the normalized drag coefficient in the BoW strategy, our proposed variants achieve within the same average accuracy range, indicating the peak performance attainable by such a strategy. Furthermore, we observed that the tokenization strategy performs better in general than the BoW strategy (see Table \ref{tbl:tok}). When this strategy adopts our proposed method, we observed the peak average accuracy reaches $95.30\%$, and 9 out of 10 candidates (or 90.00\%) in the sampled population found in the last 100th generation are practical car designs. These results demonstrate the superior performance of our proposed technique in finding viable car designs.

\begin{table}[b!]
	\centering
	\vspace{-3mm}
	\caption{Comparing between PREDO variants and its baseline in finding viable car design using the \textbf{tokenization strategy}}
        \vspace{-1mm}
	\begin{tabular}{lccc}
		\toprule
		\makecell{Design \\ Objective} & Method & \makecell{Gen. 100 \\ Avg. Accuracy} & \makecell{Overall \\ Avg. Acc.$\pm$Std. Dev.} \\
		\midrule
		Projected & Baseline & $60.00\%$ & $49.20 \pm 0.0236\%$ \\
		Frontal & $\textrm{PREDO}_{clip}$ & $80.00\%$ & $77.00 \pm 0.0274\%$ \\
        Area & $\textrm{PREDO}_{blip2}$ & $\textbf{90.00\%}$ & $\textbf{95.30} \pm \textbf{0.0049} \%$ \\
        \midrule
        Normalized & Baseline & $60.00\%$ & $64.10 \pm 0.0200 \%$ \\
		Drag & $\textrm{PREDO}_{clip}$ & $80.00\%$ & $83.30 \pm 0.0107 \%$ \\
        Coefficient & $\textrm{PREDO}_{blip2}$ & $\textbf{90.00\%}$ & $\textbf{94.60} \pm \textbf{0.0051\%}$ \\
		\bottomrule
	\end{tabular}
	\label{tbl:tok}
\end{table}

\begin{figure}[t!]
    \vspace{-2mm}
    \centering
    \footnotesize
    \begin{tabular}{c}
        \includegraphics[width=0.44\textwidth]{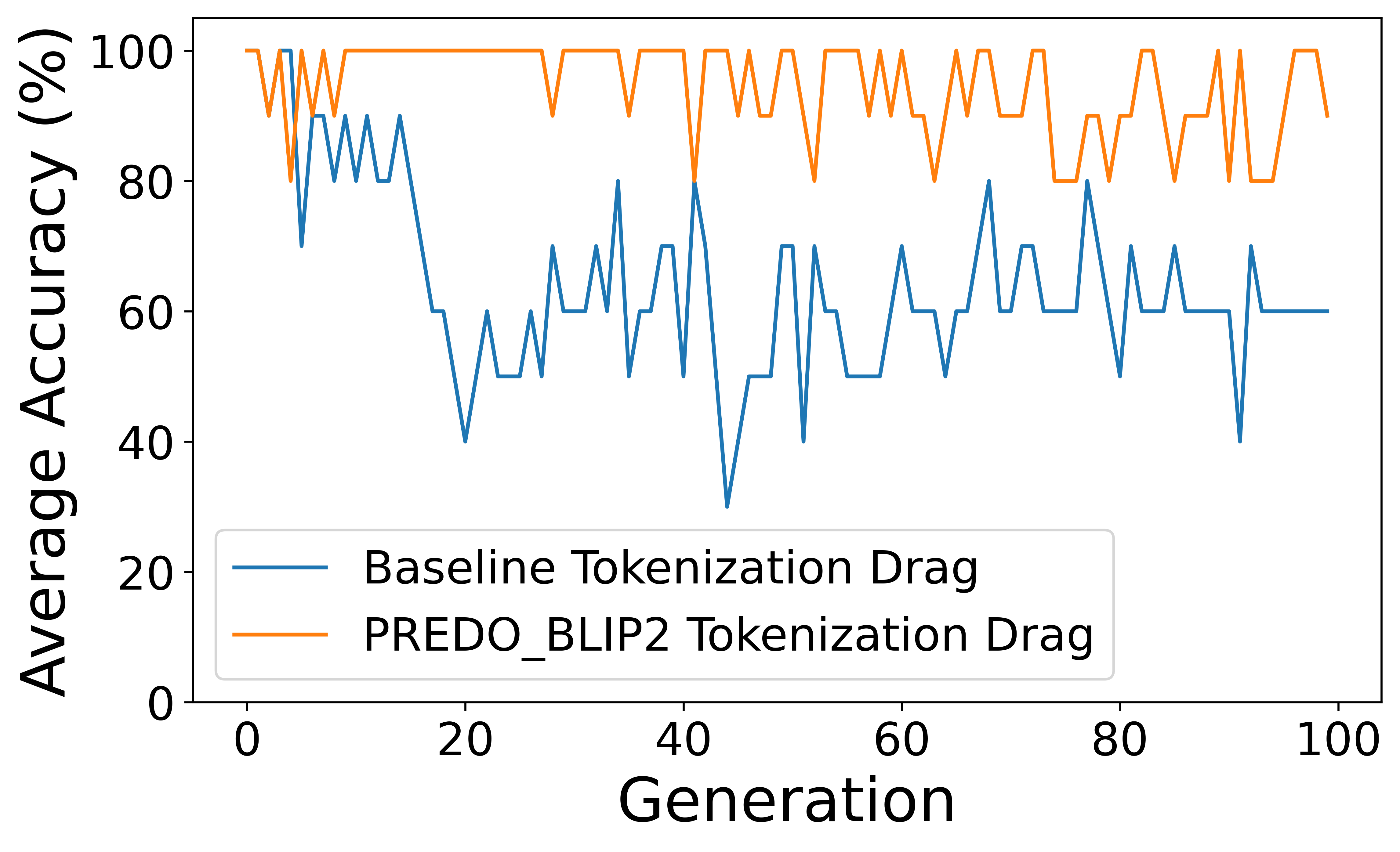} \\
        (a) Average accuracy maintained through evolution process \\[3pt]
        \includegraphics[width=0.44\textwidth]{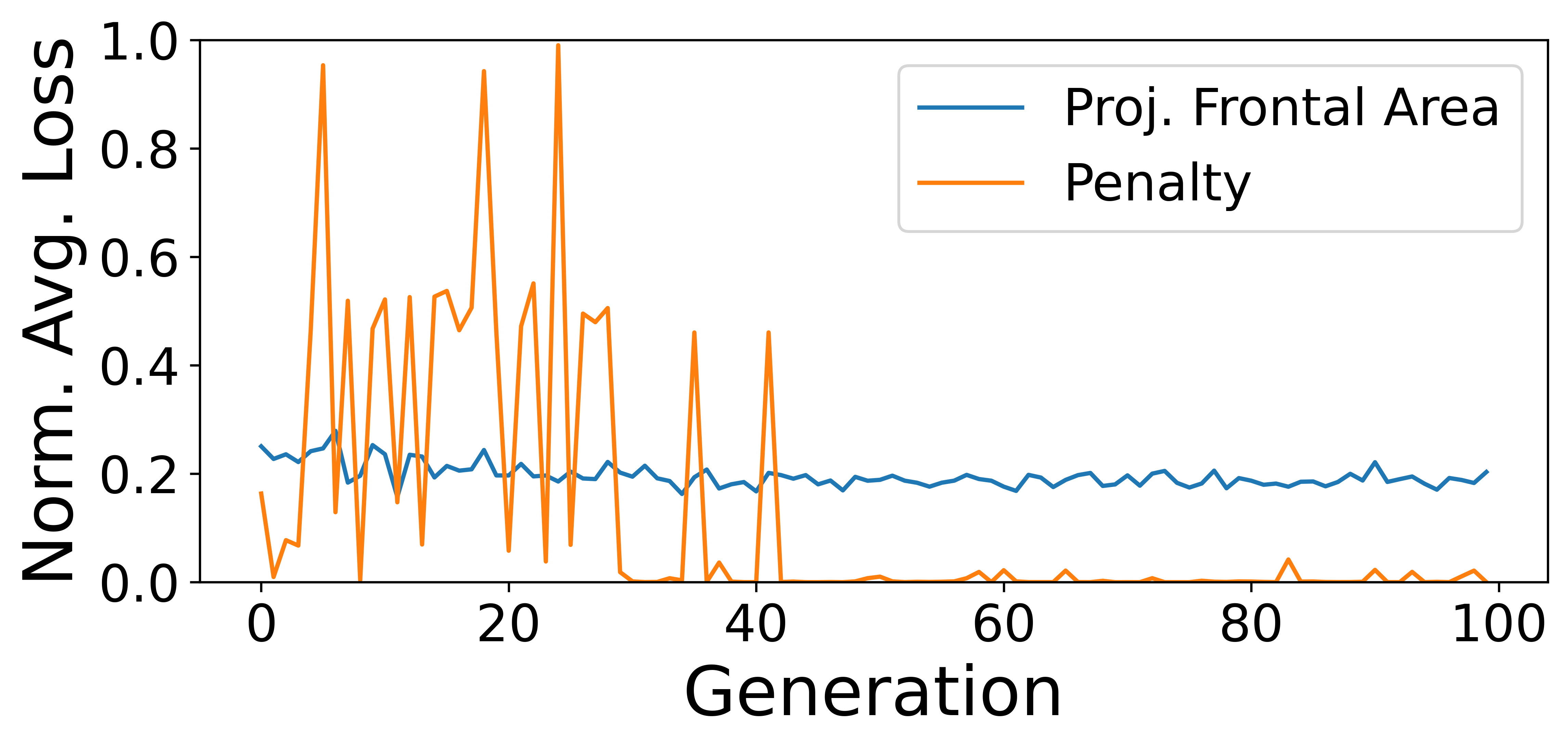} \\
        (b) Loss Components contributions during optimization \\[3pt]
        \includegraphics[width=0.44\textwidth]{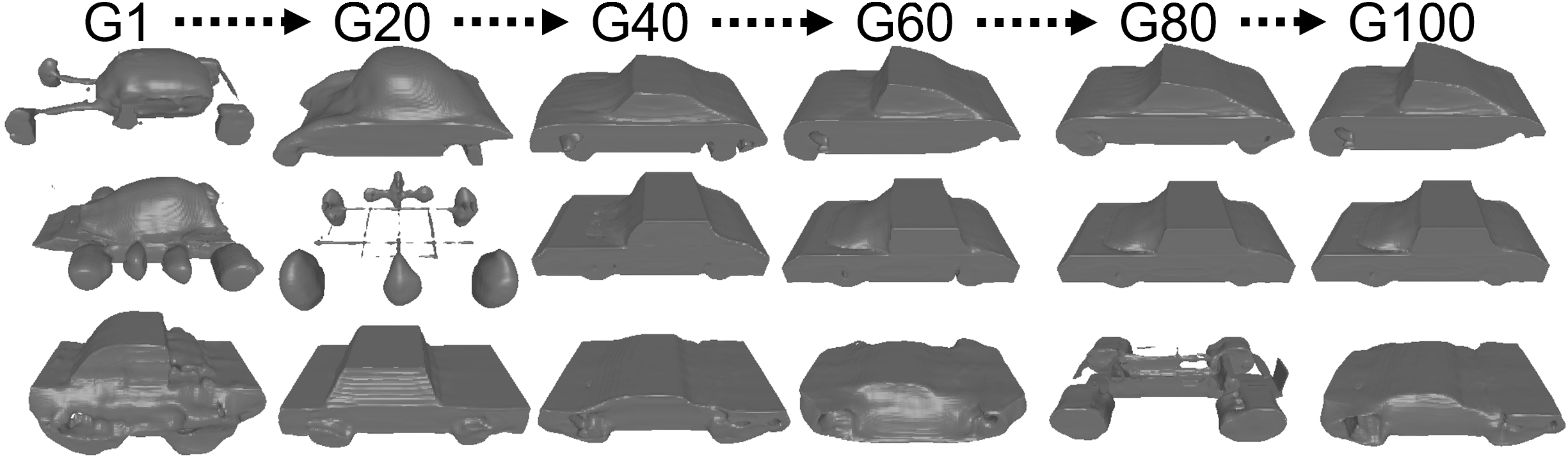} \\
        (c) Corresponding designs found during prompt evolution process \\[3pt]
    \end{tabular}
    \caption{Our proposed technique is able to maintain the majority of designs found are practical throughout the evolution process as presented in Fig. \ref{fig:accuracy}. Furthermore, the initial prompt evolution process found many impractical design candidates, as shown in Fig. 3 (b), resulting in heavy penalties imposed during this stage (Fig. \ref{fig:accuracy} (c)). As a result, the evolutionary strategy is able to move into regions having more viable designs in the later stage.}
    \label{fig:accuracy}
    \vspace{-5mm}
\end{figure}

We conducted several ablation studies to investigate the performance of our proposed technique. We observed in  Fig. \ref{fig:accuracy} (a) that using our proposed method, majority of practical designs found in each generation are viable and can maintain a high degree of accuracy throughout the generations. In contrast, the baseline method is unable to obtain such performance. This demonstrates the need for physical and visual guidance to achieve superior performance. In addition to that, we also observed that the employed text-to-3D model initially found a diverse set of novel designs where the majority are impractical car designs (see Fig. \ref{fig:accuracy}(b) and (c)), indicating the undesirable effects of ``hallucination" impacted the generative model as a consequence of evolving the text prompt to seek novel designs. This is where visual guidance is required to maintain visual preferences by imposing heavy penalties on these impractical designs during this phase. The resulting outcome effectively minimizes these ``hallucination" undesirable effects, allowing the evolution algorithm to converge toward practical designs. This demonstrates the importance of harmonizing the synchronous contribution of physics and visual guidance to facilitate the evolution algorithm in finding novel practical designs in a fully automated design evolutionary optimization framework such as ours.

\begin{figure}[t!]
    \centering
    \footnotesize
    \begin{tabular}{cc}
        \includegraphics[width=0.30\columnwidth]{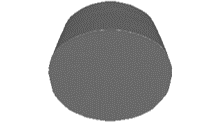} &
        \includegraphics[width=0.30\columnwidth]{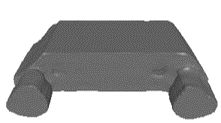} \\
        (a) Impractical Design 1 & (a) Impractical Design 2 \\[3pt]
        BLIP2 penalty=1.9355 & BLIP2 penalty=1.8894 \\[3pt]
        CLIP penalty=1.0670 & CLIP penalty=0.1468 \\[3pt]
        \includegraphics[width=0.30\columnwidth]{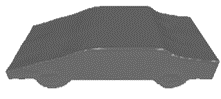} &
        \includegraphics[width=0.30\columnwidth]{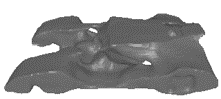} \\
        (c) Practical Design 1 & (d) Practical Design 2 \\[3pt]
        BLIP2 penalty=0.8431 & BLIP2 penalty=0.7305 \\[3pt]
        CLIP penalty=0.0000 & CLIP penalty=0.0000 \\[3pt]
    \end{tabular}
    \caption{BLIP2 vision-language model tends to assign higher penalty than CLIP across designs.}
    \label{fig:penalty}
    \vspace{-8.3mm}
\end{figure}

On the other hand, we examined the penalty imposed by CLIP and BLIP2 models. As shown in Fig. \ref{fig:penalty}, both models penalize practical designs the least as compared to impractical ones. These intended penalty outcomes allow the selection pressure to deprioritize impractical designs and encourage the selection of viable designs as parents for the next generation. Furthermore, the CLIP model exhibited the ideal scenario where no penalty is given to viable designs, conforming the designs to the physics specifications as intended. On a separate observation, we noticed that the penalty computation using Eq. \ref{eq:penalty} with the BLIP2 vision-language model tends to assign higher values than the CLIP model for the generated designs. Hence, further investigation of using the penalty weight $\alpha$ hyperparameter is necessary to study the effects of the penalty contribution. It is also worth highlighting that these vision-language models are pre-trained on different large data corpora. Hence, the prediction precision varies widely across different image-text pairs, and careful calibration is required to ensure the effectiveness of such models.

It is also worth highlighting that during our examination of the frequency and variety of tokens used in constructing the prompt in the tokenization strategy, we found that a small percentage of tokens in GPT-4 vocabulary are used in the construction, mainly concentrated in a small region. We also observed that most constructed prompts have full prompt length, much longer than the prompt used in the BoW strategy. This suggests that the prompt length directly impacts the prompt evolution performance, and further investigation in this area will be beneficial.

\begin{figure}[b!]
    \vspace{-4mm}
    \centering
    \footnotesize
    \includegraphics[width=0.45\textwidth]{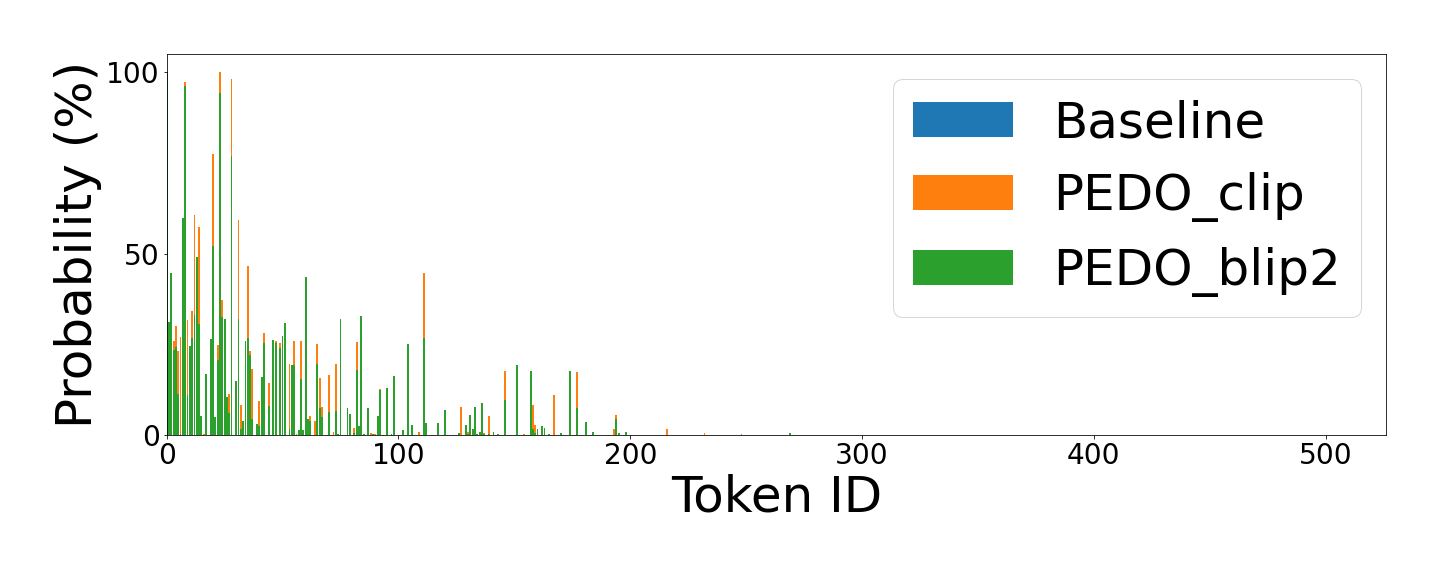} \\
    \vspace{-5mm}
    \caption{The tokenization strategy sampled the first 525 tokens in GPT-4 vocabulary during optimization to construct the text prompts, covering 0.52\% of the entire vocabulary space.}
    \label{fig:vocabfreq}
    \vspace{-0.8mm}
\end{figure}

\begin{figure*}[t!]
    \centering
    \footnotesize
    \begin{tabular}{cccc}
        \includegraphics[width=0.3\columnwidth]{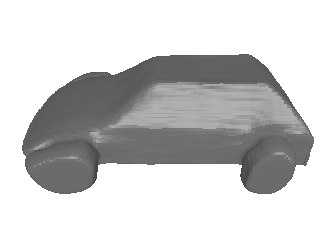} & \includegraphics[width=0.3\columnwidth]{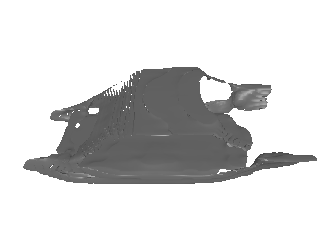} & \includegraphics[width=0.3\columnwidth]{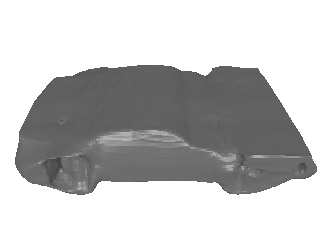} & \includegraphics[width=0.3\columnwidth]{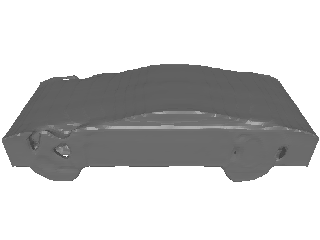} \\
        \begin{tabular}{p{6mm}p{26mm}}
    		\toprule
    	    \multicolumn{2}{c}{Best Car Design Candidate} \\
            \midrule
    		Method & Baseline \\
            \midrule
    		Obj. Score & 0.8451 \\
            \midrule
            Prompt & A occurrent car in the shape of haltere \\
    		\bottomrule
        \end{tabular} &
        \begin{tabular}{p{6mm}p{26mm}}
    		\toprule
    	    \multicolumn{2}{c}{Worst Car Design Candidate} \\
            \midrule
    		Method & Baseline \\
            \midrule
    		Obj. Score & 0.3041 \\
            \midrule
            Prompt & A occurrent car in the shape of forewing \\
    		\bottomrule
        \end{tabular} & 
        \begin{tabular}{p{6mm}p{26mm}}
    		\toprule
    	    \multicolumn{2}{c}{Best Car Design Candidate} \\
            \midrule
    		Method & $\textrm{PREDO}_{clip}$ \\
            \midrule
    		Obj. Score & 0.6529 \\
            \midrule
            Prompt & A deciding car in the shape of physical entity \\
    		\bottomrule
        \end{tabular} &
        \begin{tabular}{p{6mm}p{26mm}}
    		\toprule
    	    \multicolumn{2}{c}{Worst Car Design Candidate} \\
            \midrule
    		Method & $\textrm{PREDO}_{clip}$ \\
            \midrule
    		Obj. Score & 0.7883 \\
            \midrule
            Prompt & A knowing car in the shape of physical entity \\
    		\bottomrule
        \end{tabular} \\[5pt] \\
        \multicolumn{2}{c}{(a) Baseline, Bag-of-Words Strategy} & \multicolumn{2}{c}{(b) $\textrm{PREDO}_{clip}$, Bag-of-Words Strategy}\\[5pt]
        \includegraphics[width=0.3\columnwidth]{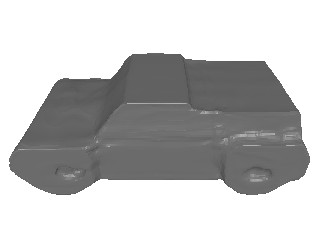} & \includegraphics[width=0.3\columnwidth]{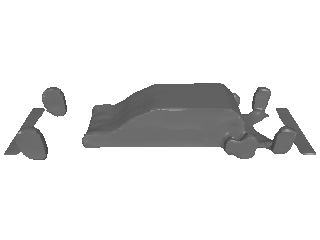} & \includegraphics[width=0.3\columnwidth]{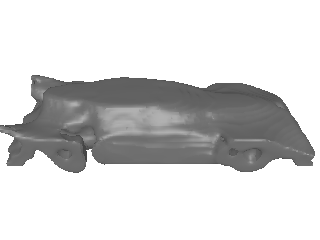} & \includegraphics[width=0.3\columnwidth]{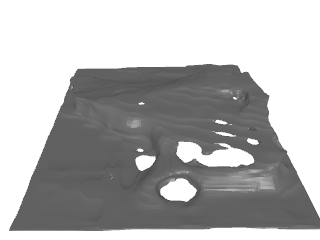} \\
        \begin{tabular}{p{6mm}p{26mm}}
    		\toprule
    	    \multicolumn{2}{c}{Best Car Design Candidate} \\
            \midrule
    		Obj. Score & 1.0376 \\
            \midrule
            Prompt & A car in the shape of R8(! reP/\}(')E!5
[*T/Cl\#5X8,=V"tsx7\%4 $>$\textbackslash K.8=@W:=9\_'0)-KH\#S)T.$>$8\%ag \\
    		\bottomrule
        \end{tabular} &
        \begin{tabular}{p{6mm}p{26mm}}
    		\toprule
    	    \multicolumn{2}{c}{Worst Car Design Candidate} \\
            \midrule
    		Obj. Score & 0.3174 \\
            \midrule
            Prompt & A car in the shape of R8(!<P/\}
            (')E!5[*T/Cl\#598,=V"ts
            8x7\%4]\textbackslash K.8=@W:=9\_'0)-KF\#S)T.>8\%: \\
    		\bottomrule
        \end{tabular} & 
        \begin{tabular}{p{6mm}p{26mm}}
    		\toprule
    	    \multicolumn{2}{c}{Best Car Design Candidate} \\
            \midrule
    		Obj. Score & 0.6595 \\
            \midrule
            Prompt & A instinct car in the shape of Lake Chelan \\
    		\bottomrule
        \end{tabular} &
        \begin{tabular}{p{6mm}p{26mm}}
    		\toprule
    	    \multicolumn{2}{c}{Worst Car Design Candidate} \\
            \midrule
    		Obj. Score & 0.3802 \\
            \midrule
            Prompt & A bent car in the shape of San Francisco Bay \\
    		\bottomrule
        \end{tabular} \\[5pt] \\
        \multicolumn{2}{c}{(c) Baseline, Tokenization Strategy} & \multicolumn{2}{c}{(d) $\textrm{PREDO}_{blip2}$, Bag-of-Words Strategy}\\[5pt]
        \includegraphics[width=0.3\columnwidth]{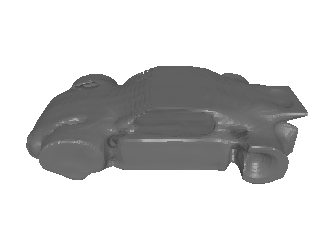} & \includegraphics[width=0.3\columnwidth]{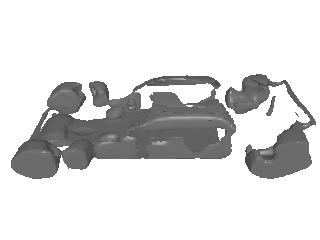} & \includegraphics[width=0.3\columnwidth]{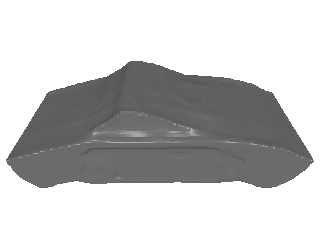} & \includegraphics[width=0.3\columnwidth]{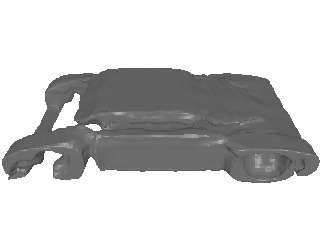} \\
        \begin{tabular}{p{6mm}p{26mm}}
    		\toprule
    	    \multicolumn{2}{c}{Best Car Design Candidate} \\
            \midrule
    		Method & Baseline \\
            \midrule
    		Obj. Score & 0.5803 \\
            \midrule
            Prompt & A car in the shape of R89!!P.\}(D)E!5\$[D*/C@ d5X8,,"sF-7*4]OM!8=@I5=j=9)'))-K\&\#b@.cV-/: \\
    		\bottomrule
        \end{tabular} &
        \begin{tabular}{p{6mm}p{26mm}}
    		\toprule
    	    \multicolumn{2}{c}{Worst Car Design Candidate} \\
            \midrule
    		Method & Baseline \\
            \midrule
    		Obj. Score & 0.3884 \\
            \midrule
            Prompt & A car in the shape of 89!!P/\}(D)E!5\$[D*/C@ \#5X8\%=,"sF-7*4]OM!8=@I5=j=9)'))-K\&\#b@.cV-/: \\
    		\bottomrule
        \end{tabular} & 
        \begin{tabular}{p{6mm}p{26mm}}
    		\toprule
    	    \multicolumn{2}{c}{Best Car Design Candidate} \\
            \midrule
    		Method & $\textrm{PREDO}_{clip}$ \\
            \midrule
    		Obj. Score & 0.7781 \\
            \midrule
            Prompt & A car in the shape of R8\#!!P.?(I)E!5\$[6d/Cl\#5" 8,=+"ts()7\%)$>$OlW!8=@\^! =u;9)x)-K)\#S$<$@.]8\%4 \\
    		\bottomrule
        \end{tabular} &
        \begin{tabular}{p{6mm}p{26mm}}
    		\toprule
    	    \multicolumn{2}{c}{Worst Car Design Candidate} \\
            \midrule
    		Method & $\textrm{PREDO}_{clip}$ \\
            \midrule
    		Obj. Score & 0.6336 \\
            \midrule
            Prompt & A car in the shape of R8\#!!P.?(u)\^!5\$Dd/C(\#5X!, =+"ts$<$()7\%)]OlW!8=@S !u;9Jx)-K@\#S$<$H.08\%4 \\
    		\bottomrule
        \end{tabular} \\[5pt] \\
        \multicolumn{2}{c}{(e) $\textrm{PREDO}_{clip}$, Tokenization Strategy} & \multicolumn{2}{c}{(f) $\textrm{PREDO}_{blip2}$, Tokenization Strategy}\\[5pt]
    \end{tabular}
    \caption{With the \textbf{projected frontal area as the design performance objective}, the worst car design candidates generated in the last 100th generation generally have lower objective scores than the corresponding best car designs. These candidates are likely to have ill-defined shapes that coincidentally lead to a smaller area than the best car designs. Our proposed technique greatly reduces the likelihood of generating a non-car-like design compared to the baseline framework.}
    \label{fig:best_vs_worst_frontal}
\end{figure*}

\begin{figure*}[t!]
    \centering
    \footnotesize
    \begin{tabular}{cccc}
        \includegraphics[width=0.3\columnwidth]{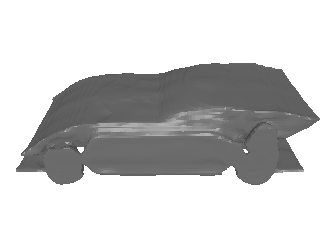} & \includegraphics[width=0.3\columnwidth]{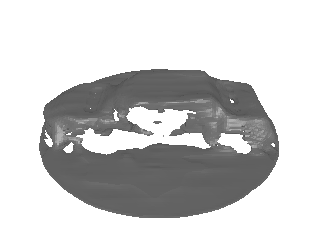} & \includegraphics[width=0.3\columnwidth]{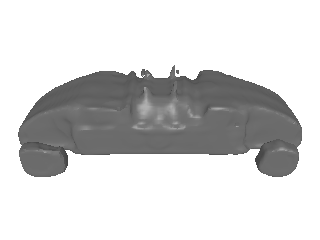} & \includegraphics[width=0.3\columnwidth]{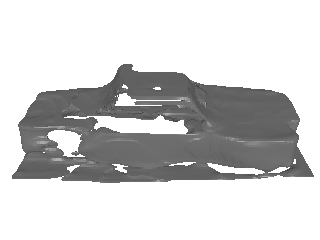} \\
        \begin{tabular}{p{6mm}p{26mm}}
    		\toprule
    	    \multicolumn{2}{c}{Best Car Design Candidate} \\
            \midrule
    		Method & Baseline \\
            \midrule
    		Obj. Score & 0.2580 \\
            \midrule
            Prompt & A stimulant car in the shape of arsenic group \\
    		\bottomrule
        \end{tabular} &
        \begin{tabular}{p{6mm}p{26mm}}
    		\toprule
    	    \multicolumn{2}{c}{Worst Car Design Candidate} \\
            \midrule
    		Method & Baseline \\
            \midrule
    		Obj. Score & 0.4634 \\
            \midrule
            Prompt & A stimulant car in the shape of Bristol Channel \\
    		\bottomrule
        \end{tabular} & 
        \begin{tabular}{p{6mm}p{26mm}}
    		\toprule
    	    \multicolumn{2}{c}{Best Car Design Candidate} \\
            \midrule
    		Method & $\textrm{PREDO}_{clip}$ \\
            \midrule
    		Obj. Score & 0.0170 \\
            \midrule
            Prompt & A deterrent car in the shape of triglyceride \\
    		\bottomrule
        \end{tabular} &
        \begin{tabular}{p{6mm}p{26mm}}
    		\toprule
    	    \multicolumn{2}{c}{Worst Car Design Candidate} \\
            \midrule
    		Method & $\textrm{PREDO}_{clip}$ \\
            \midrule
    		Obj. Score & 0.0294 \\
            \midrule
            Prompt & A unseen car in the shape of River Cam \\
    		\bottomrule
        \end{tabular} \\[5pt] \\
        \multicolumn{2}{c}{(a) Baseline, Bag-of-Words Strategy} & \multicolumn{2}{c}{(b) $\textrm{PREDO}_{clip}$, Bag-of-Words Strategy}\\[5pt]
        \includegraphics[width=0.3\columnwidth]{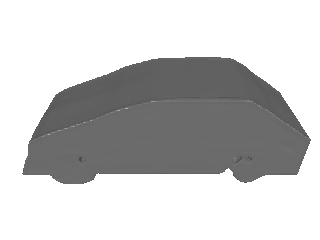} & \includegraphics[width=0.3\columnwidth]{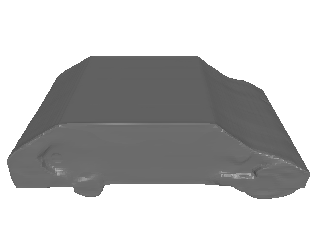} & \includegraphics[width=0.3\columnwidth]{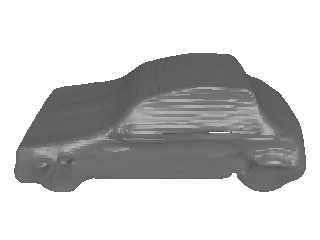} & \includegraphics[width=0.3\columnwidth]{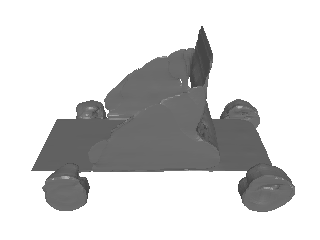} \\
        \begin{tabular}{p{6mm}p{26mm}}
    		\toprule
    	    \multicolumn{2}{c}{Best Car Design Candidate} \\
            \midrule
    		Obj. Score & 0.4995 \\
            \midrule
            Prompt & A car in the shape of RH9!!P.\}(u\^!5 [Dr/C@dX,"|s $<$ F)O\%4 \%]OK!8=BW!=9\_2E)-K\&ES)T.x8\%[ \\
    		\bottomrule
        \end{tabular} &
        \begin{tabular}{p{6mm}p{26mm}}
    		\toprule
    	    \multicolumn{2}{c}{Worst Car Design Candidate} \\
            \midrule
    		Obj. Score & 0.9417 \\
            \midrule
            Prompt & A car in the shape of RH9!!P.\}(u\^!5 \$[Dr/C@dX,"|s\$<\$F-\#\%4\%]OK!8 =BW:=9\_2E)-K\&EF@.]8\%[ \\
    		\bottomrule
        \end{tabular} & 
        \begin{tabular}{p{6mm}p{26mm}}
    		\toprule
    	    \multicolumn{2}{c}{Best Car Design Candidate} \\
            \midrule
    		Obj. Score & 0.0239 \\
            \midrule
            Prompt & A rallying car in the shape of nonessential \\
    		\bottomrule
        \end{tabular} &
        \begin{tabular}{p{6mm}p{26mm}}
    		\toprule
    	    \multicolumn{2}{c}{Worst Car Design Candidate} \\
            \midrule
    		Obj. Score & 0.1309 \\
            \midrule
            Prompt & A spanking car in the shape of dander \\
    		\bottomrule
        \end{tabular} \\[5pt] \\
        \multicolumn{2}{c}{(c) Baseline, Tokenization Strategy} & \multicolumn{2}{c}{(d) $\textrm{PREDO}_{blip2}$, Bag-of-Words Strategy}\\[5pt]
        \includegraphics[width=0.3\columnwidth]{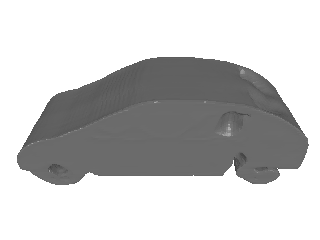} & \includegraphics[width=0.3\columnwidth]{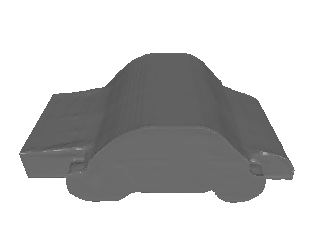} & \includegraphics[width=0.3\columnwidth]{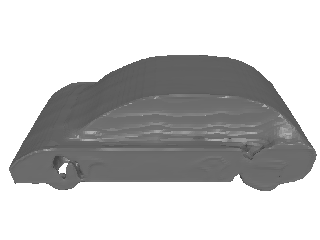} & \includegraphics[width=0.3\columnwidth]{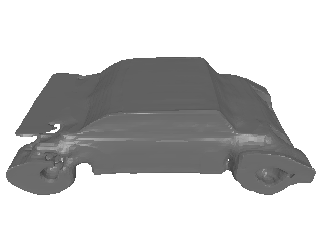} \\
        \begin{tabular}{p{6mm}p{26mm}}
    		\toprule
    	    \multicolumn{2}{c}{Best Car Design Candidate} \\
            \midrule
    		Method & Baseline \\
            \midrule
    		Obj. Score & 0.5145 \\
            \midrule
            Prompt & A car in the shape of RH(!!P/?(')E$<$5 \$[D|C@\#5X,=,"[sV-O*z]OKi=@FI:=j9.2))K\&\#b4.K8\%: \\
    		\bottomrule
        \end{tabular} &
        \begin{tabular}{p{6mm}p{26mm}}
    		\toprule
    	    \multicolumn{2}{c}{Worst Car Design Candidate} \\
            \midrule
    		Method & Baseline \\
            \midrule
    		Obj. Score & 1.090 \\
            \midrule
            Prompt & A car in the shape of RH9!!w/?(')E$<$5 \$[D|C@\#5X,=,"[sV-O*z]OKi=@FI:=j9.2))K\&\#b4.x8\%: \\
    		\bottomrule
        \end{tabular} & 
        \begin{tabular}{p{6mm}p{26mm}}
    		\toprule
    	    \multicolumn{2}{c}{Best Car Design Candidate} \\
            \midrule
    		Method & $\textrm{PREDO}_{clip}$ \\
            \midrule
    		Obj. Score & 0.6050 \\
            \midrule
            Prompt & A car in the shape of R89!!P.?(I)EA5[DT/Cu\# 5X,=+"tp$<$(-\}*-U\&Hlg8=BnW:= u.9J'))CK)\#)4.08/: \\
    		\bottomrule
        \end{tabular} &
        \begin{tabular}{p{6mm}p{26mm}}
    		\toprule
    	    \multicolumn{2}{c}{Worst Car Design Candidate} \\
            \midrule
    		Method & $\textrm{PREDO}_{clip}$ \\
            \midrule
    		Obj. Score & 1.1094 \\
            \midrule
            Prompt & A car in the shape of R89!!P.?(I)EA5[DT/Cu\# 5X,=+"tp\$<\$(-\}*-U\&Hlg8=BkW:= u.96'))CK)\#)4.08/: \\
    		\bottomrule
        \end{tabular} \\[5pt] \\
        \multicolumn{2}{c}{(e) $\textrm{PREDO}_{clip}$, Tokenization Strategy} & \multicolumn{2}{c}{(f) $\textrm{PREDO}_{blip2}$, Tokenization Strategy}\\[5pt]
    \end{tabular}
    \caption{\textbf{Normalized drag coefficient as the design performance objective} has the better association between the geometric structure and the text prompt used to synthesize the designs, yielding the best design candidates with the least objective scores as observed in the terminating 100th generation.}
    \label{fig:best_vs_worst_drag}
\end{figure*}

\begin{figure*}[t!]
    \centering
    \footnotesize
    \begin{tabular}{cc}
     \includegraphics[width=0.5\textwidth]{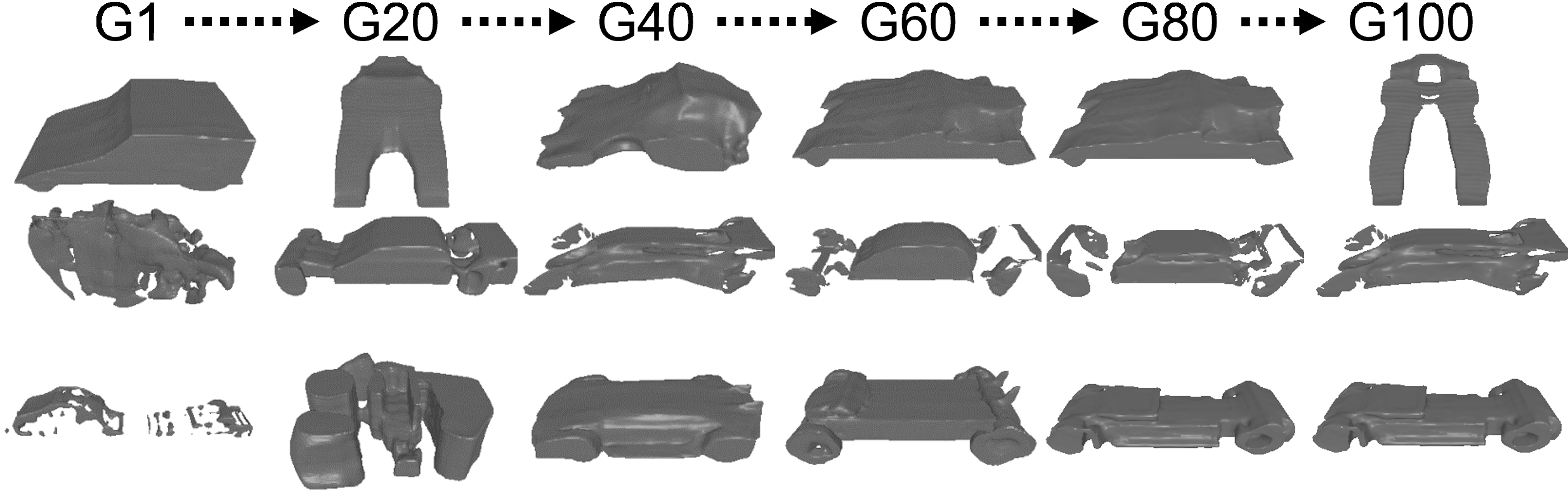} & \includegraphics[width=0.5\textwidth]{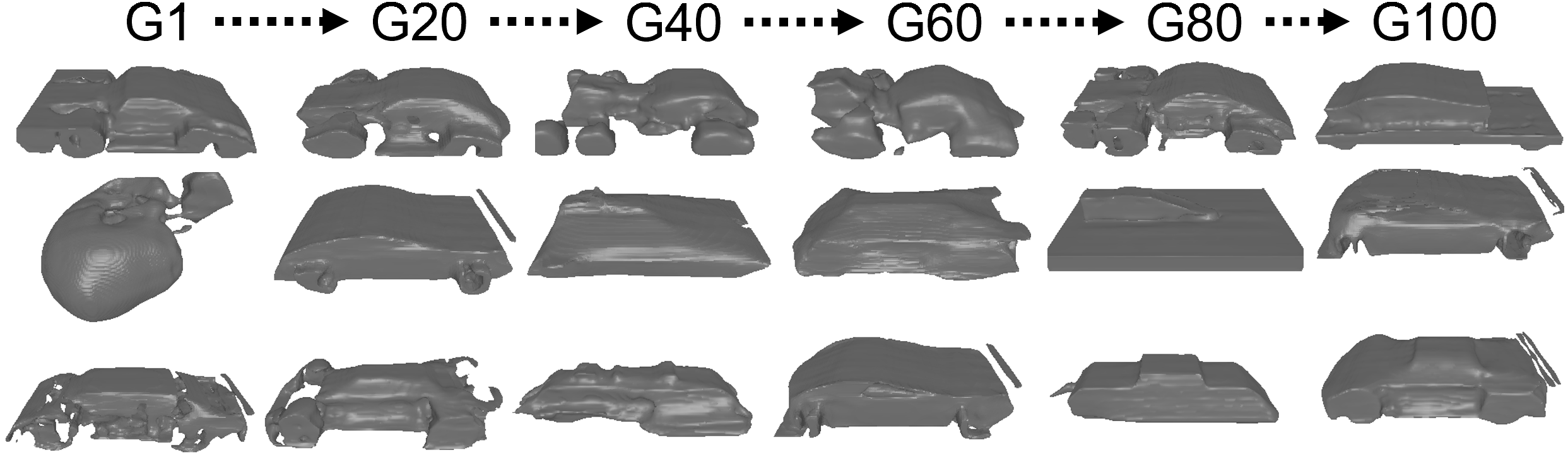} \\
     (a) Baseline, Bag-of-Words Strategy, Projected Frontal Area & (b) Baseline, Bag-of-Words Strategy, Normalized Drag Coefficient \\[5pt]
     \includegraphics[width=0.5\textwidth]{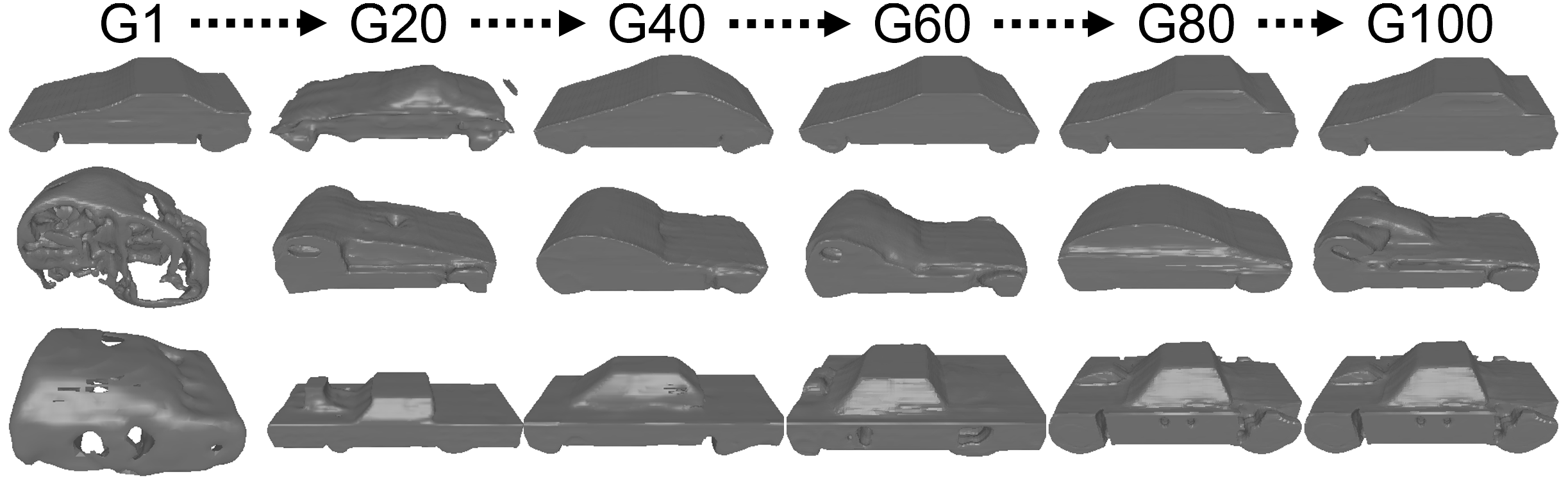} & \includegraphics[width=0.5\textwidth]{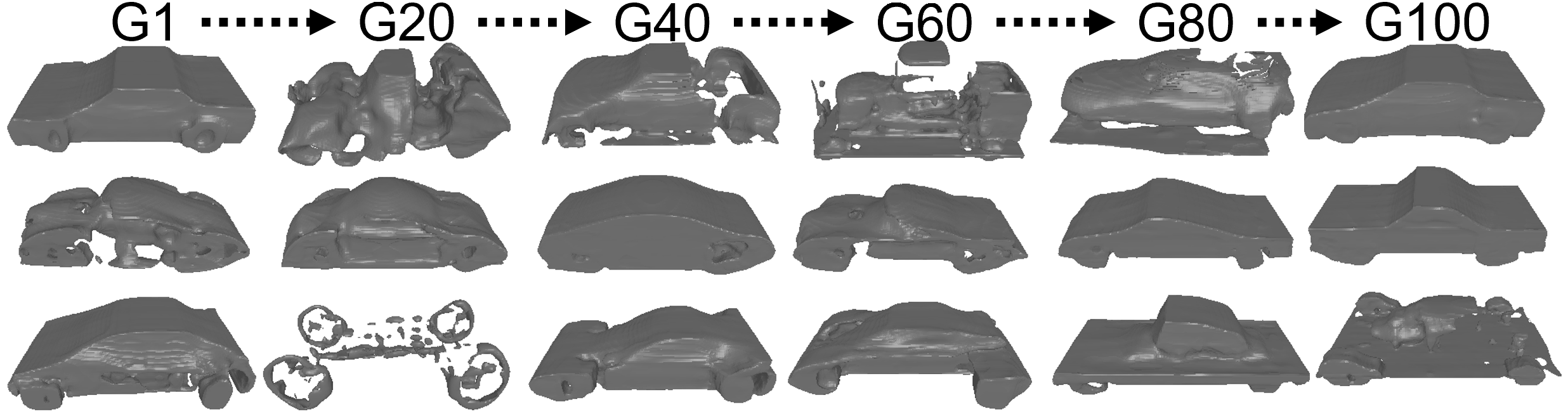} \\
     (c) $\textrm{PREDO}_{clip}$, Bag-of-Words Strategy, Projected Frontal Area & (d) $\textrm{PREDO}_{clip}$, Bag-of-Words Strategy, Normalized Drag Coefficient \\[5pt]
     \includegraphics[width=0.5\textwidth]{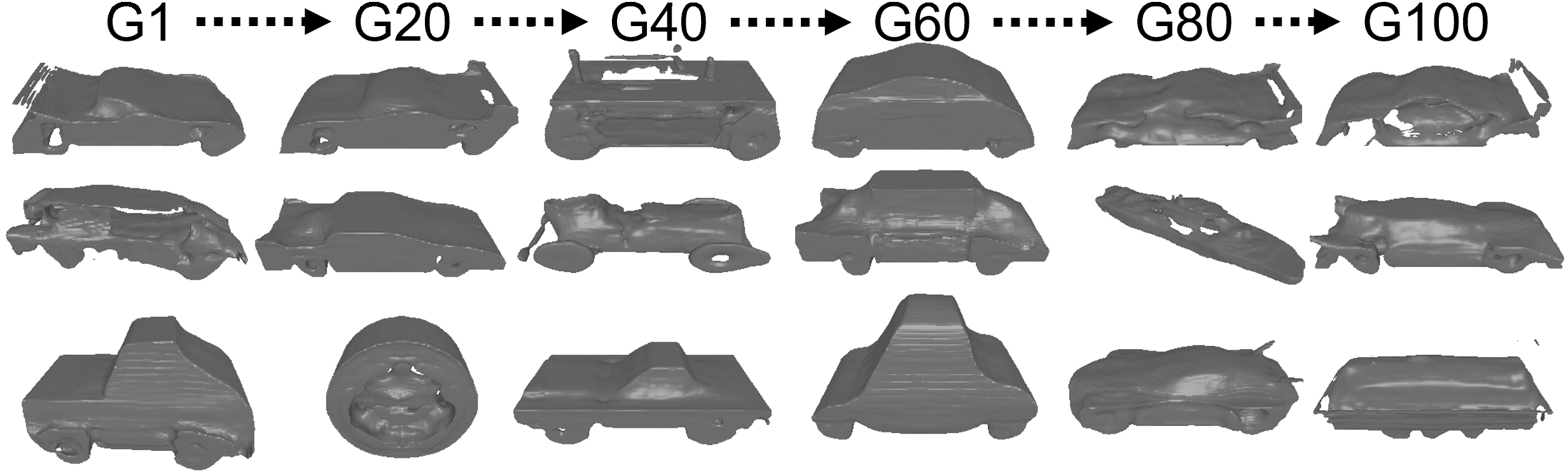} & \includegraphics[width=0.5\textwidth]{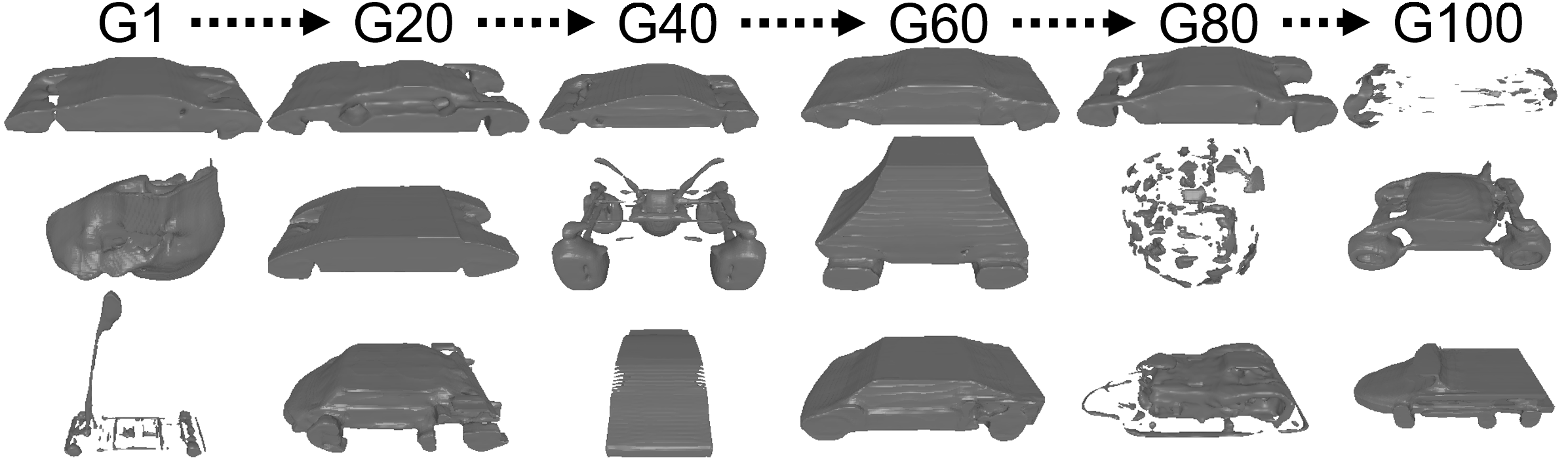} \\
     (e) $\textrm{PREDO}_{blip2}$, Bag-of-Words Strategy, Projected Frontal Area & (f) $\textrm{PREDO}_{blip2}$, Bag-of-Words Strategy, Normalized Drag Coefficien \\[5pt]
     \includegraphics[width=0.5\textwidth]{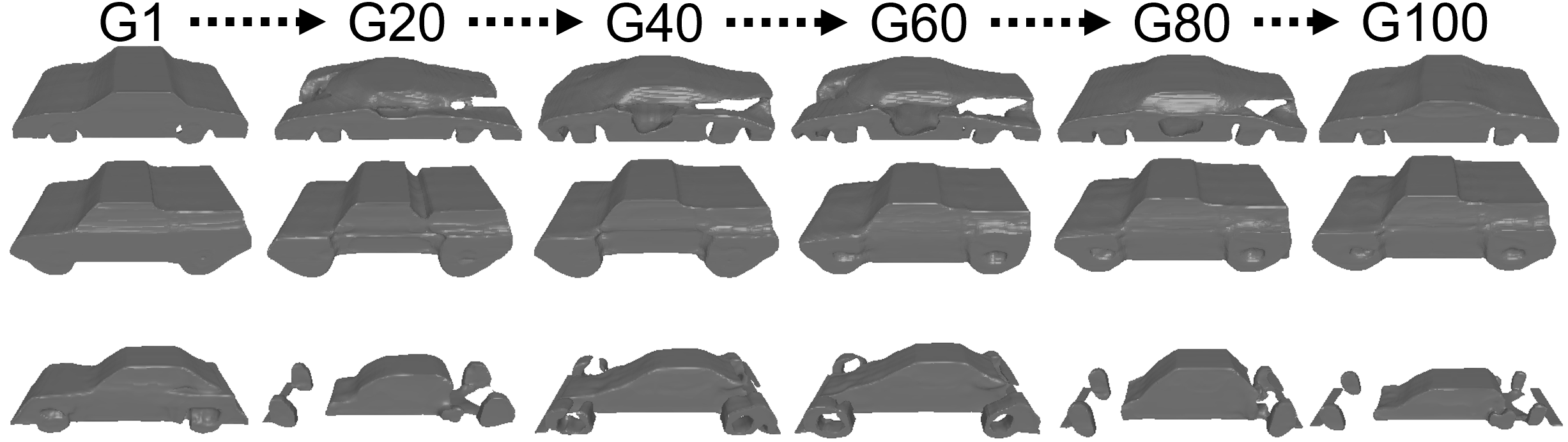} & \includegraphics[width=0.5\textwidth]{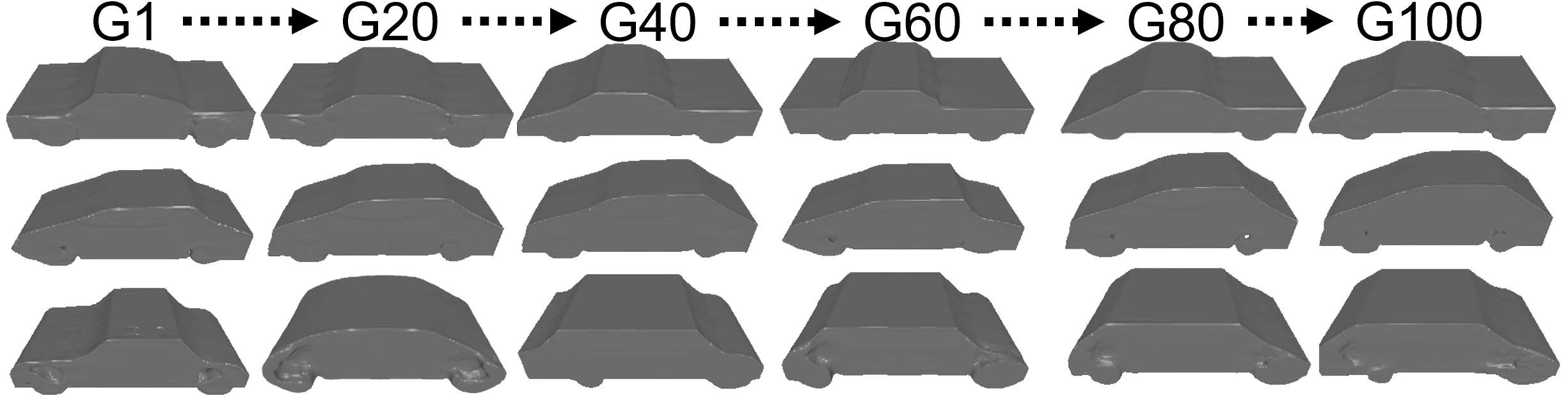} \\
     (g) Baseline, Tokenization Strategy, Projected Frontal Area & (h) Baseline, Tokenization Strategy, Normalized Drag Coefficient \\[5pt]
     \includegraphics[width=0.5\textwidth]{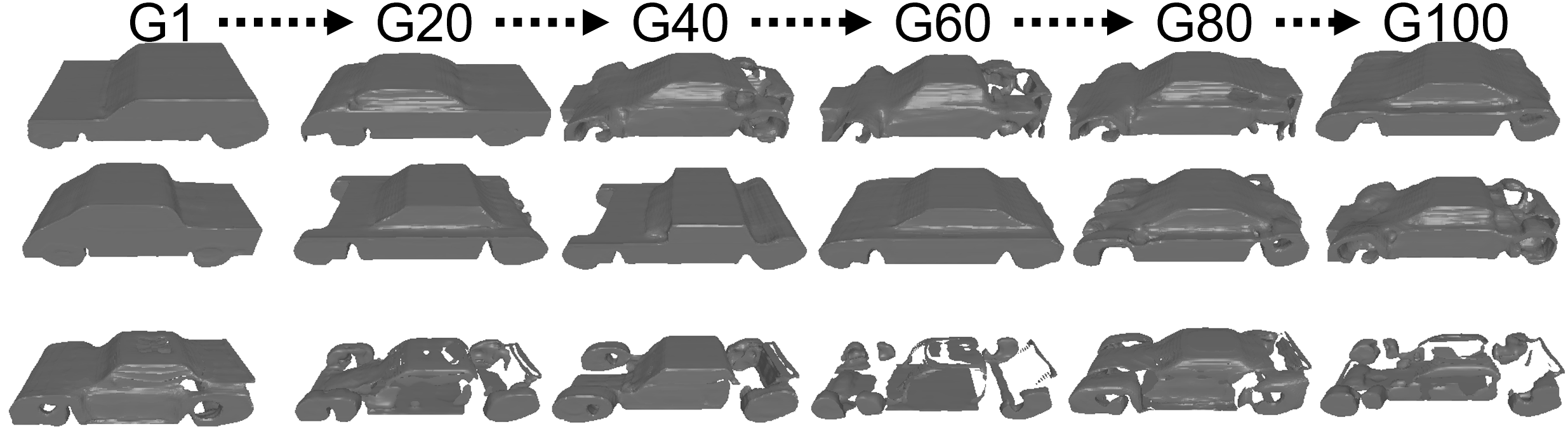} & \includegraphics[width=0.5\textwidth]{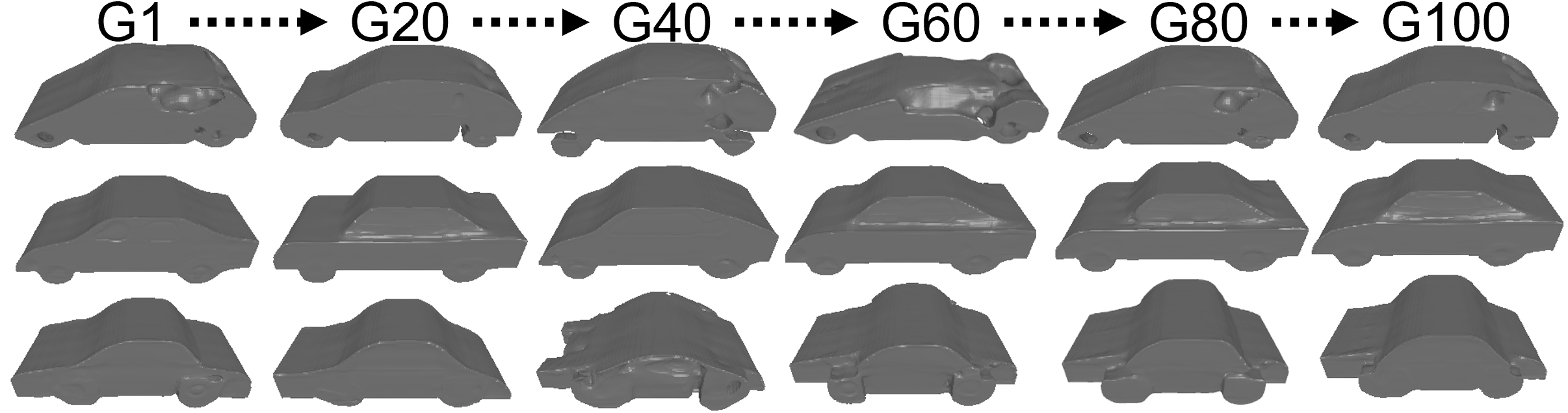} \\
     (i) $\textrm{PREDO}_{clip}$, Tokenization Strategy, Projected Frontal Area & (j) $\textrm{PREDO}_{clip}$, Tokenization Strategy, Normalized Drag Coefficientt \\[5pt]
     \includegraphics[width=0.5\textwidth]{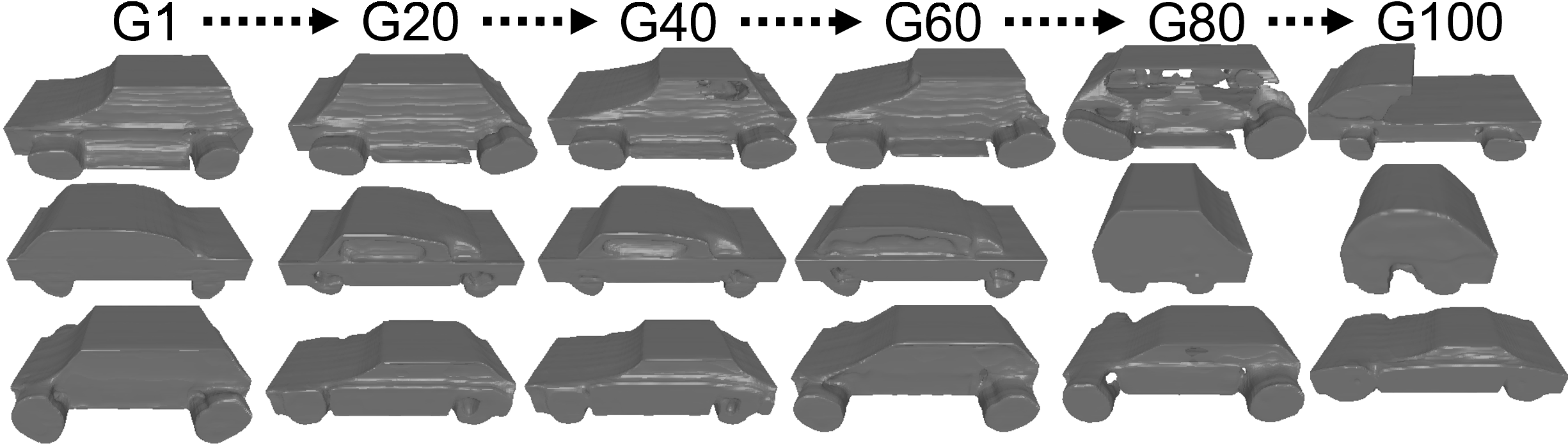} & \includegraphics[width=0.5\textwidth]{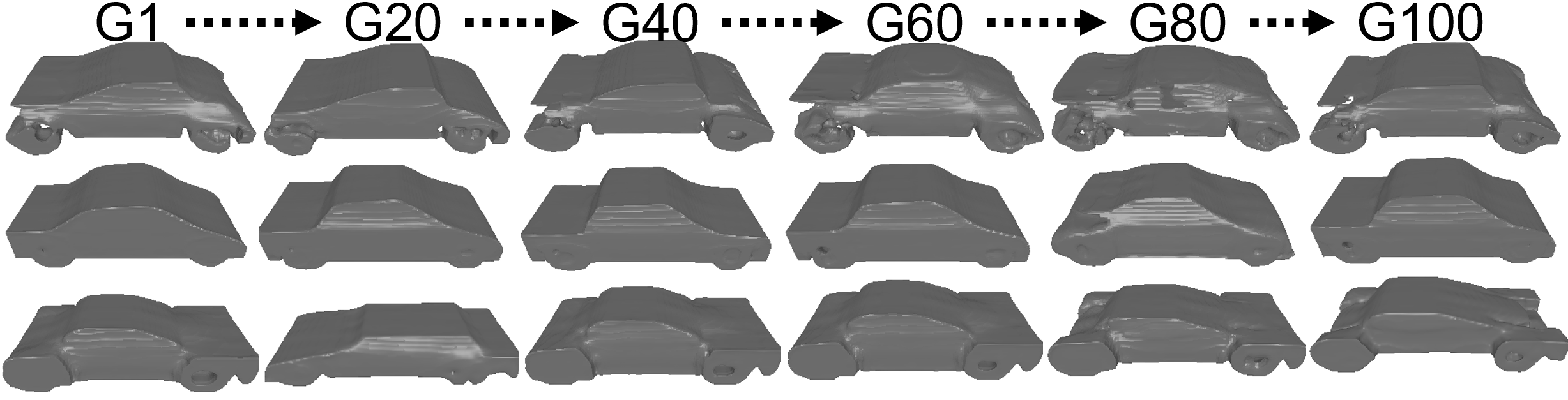} \\
     (k) $\textrm{PREDO}_{blip2}$, Tokenization Strategy, Projected Frontal Area & (l) $\textrm{PREDO}_{blip2}$, Tokenization Strategy, Normalized Drag Coefficient \\[5pt]
    \end{tabular}
    \caption{Car design candidates generated by text-to-3D model during the prompt evolution process, from generation 1 (G1) to maximum generation 100 (G100), have demonstrated that our proposed technique facilitates the natural evolution of car geometric structure, especially to the tokenization strategy with normalized drag coefficient as design performance (see Fig. \ref{fig:evo} (h), Fig. \ref{fig:evo} (j) and Fig. \ref{fig:evo} (l)), leading to more car-like designs to be found during the prompt evolution process.}
    \label{fig:evo}
\end{figure*}

Last but not least, we conducted visual inspections of the generated car designs across the generations and compared them against the performance results presented in Table \ref{tbl:bow} and Table \ref{tbl:tok}. 
Interestingly, we observed design candidates that yield the least projected car frontal area design performance score generally are, in fact, ill-defined designs. Our proposed technique that uses a vision-language model can minimize the likelihood of the derivative-free optimizer finding such ill-defined designs as shown in Fig. \ref{fig:best_vs_worst_frontal} (b), (c) and (f). In contrast, as illustrated in Fig. \ref{fig:best_vs_worst_drag}, using the normalized drag coefficient as the design performance yields the best car design candidates and indicates a better association between the geometric structure of the car and the text prompt used to generate these designs. We also examined the designs found during the prompt evolution process in Fig. \ref{fig:evo} and noticed that once the framework can disincentivize substantial ill-defined designs, the evolution of the geometric structure is more natural, resulting in more car-like designs found (see Fig. \ref{fig:evo} (h), Fig. \ref{fig:evo} (j) and Fig. \ref{fig:evo} (l)). 

The experimental results and analysis presented in this paper highlight that prompt evolution is a promising paradigm for finding novel designs with a high degree of accuracy maintained throughout the evolutionary process while providing ease of use in specifying design specifications and preferences via a natural language interface. Highlighted in \cite{wong2023promptevo}, prompt evolution can exert evolutionary selection pressure and variation to the generated outputs and other forms of guidance and hidden states. In this paper, we have demonstrated this possibility in the form of text prompts. Given such flexibility within this emerging paradigm, prompt evolution presents abundant research opportunities in generative AI for evolutionary engineering design optimization.

\vspace{-2mm}
\section{Conclusion and Outlook} \label{sec:conclusion}
\vspace{-1mm}

In this paper, we proposed a novel evolutionary design optimization framework (PREDO) that employs the prompt evolution paradigm and incorporates a vision-language model to penalize impractical designs and rank practical ones higher than others. We also introduced a design optimization objective that employs a physics solver and a vision-language model for practical or functional guidance in the generated designs. With these contributions, we address the challenge faced in an earlier study \cite{rios2023LLM} that has ill-defined generated designs served as parents for the next-generation offspring, resulting in reduced performance of the framework.

Our experiment results and ablation studies demonstrated the effectiveness of our proposed technique and the superior performance achieved in the presented car design optimization example. This solidifies the possibility of using the prompt evolution paradigm in a fully automated evolutionary design optimization framework as an alternative to traditional engineering design optimization. 

Nevertheless, further investigations into the prompt evolution paradigm and the framework are necessary to fully exploit the capability of evolutionary algorithms in such a setting. One particular area highlighted is to study the impact of prompt length on the prompt evolution performance. In addition, it is necessary to investigate the effects of the penalty contribution in relation to different penalty weights. Another area of investigation is the exploration of the proposed methods to other types of engineering design applications, \textit{e.g.} with respect to other aero-/fluiddynamic domains (planes, boats) or computationally-costly physics modalities as structural optimization including finite-element solvers. Lastly, a noteworthy research area is to study the relationship between the geometric difference and textual descriptions in the prompt on prompt evolution performance.

\vspace{-2mm}
\section*{Acknowledgment}
\vspace{-1mm}

The authors would like to express our deep gratitude to individuals, especially Dr. Jiao Liu, for their expertise and valuable critiques in this research work. This research is partly supported by the Honda Research Institute Europe (HRIEU) and the College of Computing \& Data Science (CCDS), Nanyang Technological University (NTU). 

\bibliographystyle{IEEEtran}
\vspace{-2mm}
\bibliography{References}
\vspace{-1mm}

\end{document}